\documentclass[10pt,twocolumn,letterpaper]{article}

\usepackage{cvpr}
\usepackage{times}
\usepackage{graphicx}
\usepackage{amsmath}
\usepackage{amssymb}

\usepackage{mathtools}
\usepackage{xcolor}
\usepackage{multirow}
\usepackage{hhline}

\usepackage[pagebackref=true,breaklinks=true,letterpaper=true,colorlinks,bookmarks=false]{hyperref}

\cvprfinalcopy 

\def\cvprPaperID{1509} 

\ifcvprfinal\fi
\begin{document}
	
	\title{Deep Multi-scale Convolutional Neural Network for Dynamic Scene Deblurring}
	
	\author{Seungjun Nah \and Tae Hyun Kim \and Kyoung Mu Lee \and \\
		Department of ECE, ASRI, Seoul National University, 08826, Seoul, Korea\\
		{\tt\small \{seungjun.nah, lliger9\}@gmail.com, kyoungmu@snu.ac.kr}
	}

	\maketitle

	\begin{abstract}
		
		Non-uniform blind deblurring for general dynamic scenes is a challenging computer vision problem as blurs arise not only from multiple object motions but also from camera shake, scene depth variation. To remove these complicated motion blurs, conventional energy optimization based methods rely on simple assumptions such that blur kernel is partially uniform or locally linear. Moreover, recent machine learning based methods also depend on synthetic blur datasets generated under these assumptions. This makes conventional deblurring methods fail to remove blurs where blur kernel is difficult to approximate or parameterize (e.g. object motion boundaries).
		In this work, we propose a multi-scale convolutional neural network that restores sharp images in an end-to-end manner where blur is caused by various sources. Together, we present multi-scale loss function that mimics conventional coarse-to-fine approaches. Furthermore, we propose a new large-scale dataset that provides pairs of realistic blurry image and the corresponding ground truth sharp image that are obtained by a high-speed camera.
		With the proposed model trained on this dataset, we demonstrate empirically that our method achieves the state-of-the-art performance in dynamic scene deblurring not only qualitatively, but also quantitatively.
		
	\end{abstract}
	
	\section{Introduction}
	
	%
	%

	Motion blur is one of the most commonly arising types of artifacts when taking photos. Shakes of camera and fast object motions degrade image quality to undesired blurry images.
	Furthermore, various causes such as depth variation, occlusion in motion boundaries make blurs even more complex.
	Single image deblurring problem is to estimate the unknown sharp image given a blurry image.
	Earlier studies focused on removing blurs caused by simple translational or rotational camera motions. More recent works try to handle general non-uniform blurs caused by depth variation, camera shakes and object motions in dynamic environments. Most of these approaches are based on following blur model ~\cite{whyte2010non,gupta2010single,hirsch2011fast,harmeling2010space}.
	
	\begin{equation}
	\textbf{B} = \textbf{K}\textbf{S}+\textbf{n},
	\label{equ_blur_constraint}
	\end{equation}
	
	where \textbf{B}, \textbf{S} and \textbf{n} are vectorized blurry image, latent sharp image, and noise, respectively. \textbf{K} is a large sparse matrix whose rows each contain a local blur kernel acting on \textbf{S} to generate a blurry pixel.
	In practice, blur kernel is unknown. Thus, blind deblurring methods try to estimate latent sharp image $\textbf{S}$ and blur kernel $\textbf{K}$ simultaneously.
	
	Finding blur kernel for every pixel is a severely ill-posed problem. Thus, some approaches tried to parametrize blur models with simple assumptions on the sources of blurs. In ~\cite{whyte2010non, gupta2010single}, they assumed that blur is caused by 3D camera motion only. However, in dynamic scenes, the kernel estimation is more challenging as there are multiple moving objects as well as camera motion. Thus, Kim et al.~\cite{thkim:2013} proposed a dynamic scene deblurring method that jointly segments and deblurs a non-uniformly blurred image, allowing the estimation of complex (non-linear) kernel within a segment. In addition, Kim and Lee~\cite{thkim:2014} approximated the blur kernel to be locally linear and proposed an approach that estimates both the latent image and the locally linear motions jointly.
	However, these blur kernel approximations are still inaccurate, especially in the cases of abrupt motion discontinuities and occlusions. Note that such erroneous kernel estimation directly affects the quality of the latent image, resulting in undesired ringing artifacts.
	
	\begin{figure*}[t]
		\begin{center}
			\includegraphics[width=\linewidth]{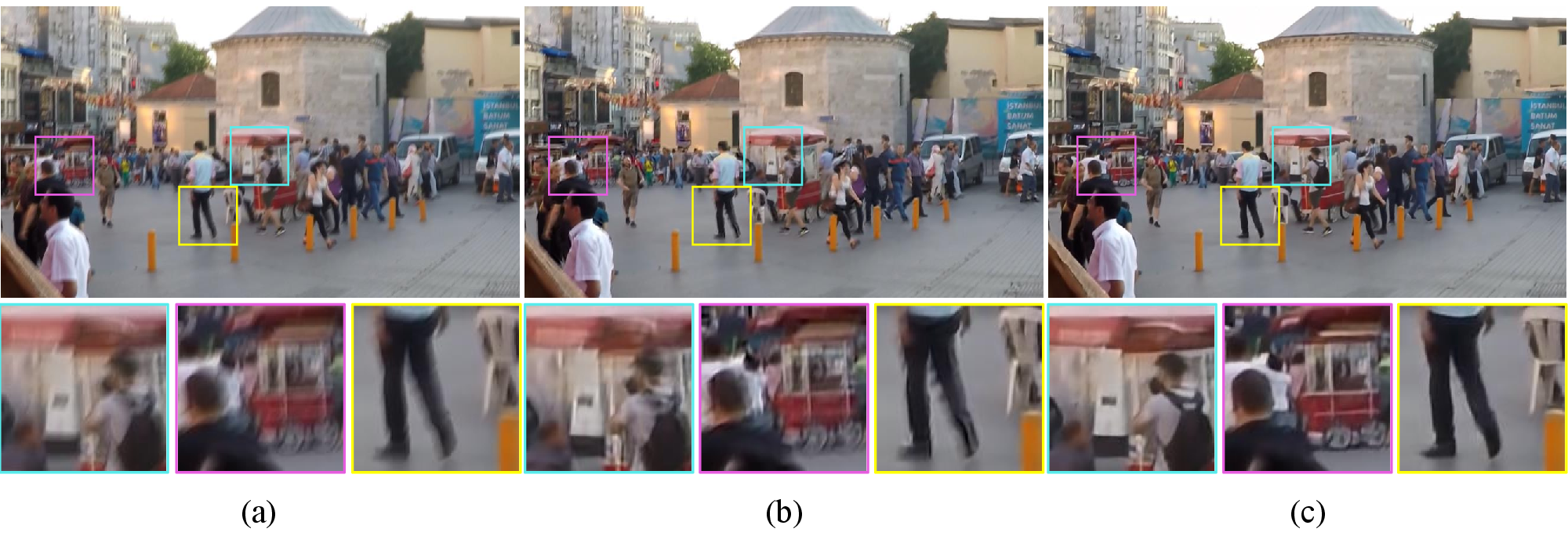}
		\end{center}
		\caption{(a) Input blurry image. (b) Result of Sun et al.~\cite{sun2015learning}. (c) Our deblurring result. 
			Our results show clear object boundaries without artifacts.}
		\label{fig_intro}
	\end{figure*}

	Recently, CNNs (Convolutional Neural Networks) have been applied in numerous computer vision problems including deblurring problem and showed promising results ~\cite{xu2014deep, schuler2016learning, sun2015learning, chakrabarti2016neural}. Since no pairs of real blurry image and ground truth sharp image are available for supervised learning, they commonly used blurry images generated by convolving synthetic blur kernels. In ~\cite{xu2014deep,schuler2016learning,chakrabarti2016neural}, synthesized blur images with uniform blur kernel are used for training. And, in ~\cite{sun2015learning}, classification CNN is trained to estimate locally linear blur kernels. Thus, CNN-based models are still suited only to some specific types of blurs, and there are restrictions on more common spatially varying blurs.
	
	Therefore, all the existing methods still have many problems before they could be generalized and used in practice. These are mainly due to the use of simple and unrealistic blur kernel models. Thus, to solve those problems, in this work, we propose a novel end-to-end deep learning approach for dynamic scene deblurring.
	
	First, we propose a multi-scale CNN that directly restores latent images without assuming any restricted blur kernel model. Especially, the multi-scale architecture is designed to mimic conventional coarse-to-fine optimization methods. Unlike other approaches, our method does not estimate explicit blur kernels. Accordingly, our method is free from artifacts that arise from kernel estimation errors. 
	Second, we train the proposed model with a multi-scale loss that is appropriate for coarse-to-fine architecture that enhances convergence greatly. In addition, we further improve the results by employing adversarial loss~\cite{goodfellow2014generative}.
	Third, we propose a new realistic blurry image dataset with ground truth sharp images.
	To obtain kernel model-free dataset for training, we employ the dataset acquisition method introduced in~\cite{kim2016dynamic}.
	As the blurring process can be modeled by the integration of sharp images during shutter time~\cite{kim2016dynamic,li2010generating,thkim:2015}, we captured a sequence of sharp frames of a dynamic scene with a high-speed camera and averaged them to generate a blurry image by considering gamma correction.
	
	By training with the proposed dataset and adding proper augmentation, our model can handle general local blur kernel implicitly. As the loss term optimizes the result to resemble the ground truth,
	it even restores occluded regions where blur kernel is extremely complex as shown in Fig.~\ref{fig_intro}.
	We trained our model with millions of pairs of image patches and achieved significant improvements in dynamic scene deblurring. Extensive experimental results demonstrate that the performance of the proposed method is far superior to those of the state-of-the-art dynamic scene deblurring methods in both qualitative and quantitative evaluations.

	\subsection{Related Works}
	
	There are several approaches that employed CNNs for deblurring~\cite{xu2014deep, sun2015learning, schuler2016learning, chakrabarti2016neural}.
	
	Xu et al.~\cite{xu2014deep} proposed an image deconvolution CNN to deblur a blurry image in a non-blind setting. They built a network based on the separable kernel property that the (inverse) blur kernel can be decomposed into a small number of significant filters.
	Additionally, they incorporated the denoising network~\cite{eigen2013restoring} to reduce visual artifacts such as noise and color saturation by concatenating the module at the end of their proposed network.
	
	On the other hand, Schuler et al.~\cite{schuler2016learning} proposed a blind deblurring method with CNN. Their proposed network mimics conventional optimization-based deblurring methods and iterates the feature extraction, kernel estimation, and the latent image estimation steps in a coarse-to-fine manner. To obtain pairs of sharp and blurry images for network training, they generated uniform blur kernels using a Gaussian process and synthesized lots of blurry images by convolving them to the sharp images collected from the ImageNet dataset~\cite{deng2009imagenet}. However, they reported performance limits for large blurs due to their suboptimal architecture.
	
	Similarly to the work of Couzinie-Devy et al.~\cite{couzinie2013learning}, Sun et al.~\cite{sun2015learning} proposed a sequential deblurring approach.
	First, they generated pairs of blurry and sharp patches with 73 candidate blur kernels.
	Next, they trained classification CNN to measure the likelihood of a specific blur kernel of a local patch. And then smoothly varying blur kernel is obtained by optimizing an energy model that is composed of the CNN likelihoods and smoothness priors. Final latent image estimation is performed with conventional optimization method~\cite{zoran2011learning}.
	
	Note that all these methods require an accurate kernel estimation step for restoring the latent sharp image. In contrast, our proposed model is learned to produce the latent image directly without estimating blur kernels.
	
	\begin{figure*}[t]
		\begin{center}
			\includegraphics[width=\linewidth]{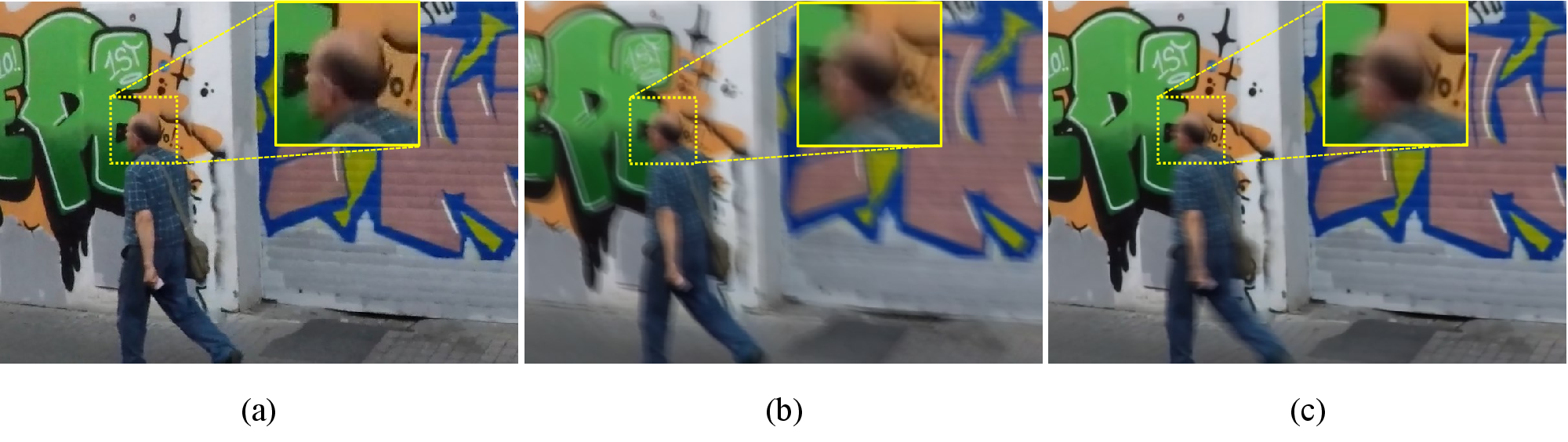}
		\end{center}
		\caption{(a) Ground truth sharp image. (b) Blurry image generated by convolving a uniform blur kernel. (c) Blurry image by averaging sharp frames.
			In this case, blur is mostly caused by person motion, leaving the background as it is. The blur kernel is non-uniform, complex shaped. However, when the blurry image is synthesized by convolution with a uniform kernel, the background also gets blurred as if blur was caused by camera shake. To model dynamic scene blur, our kernel-free method is required.
		}
		\label{fig_blur_comp}
	\end{figure*}
	
	In other computer vision tasks, several forms of coarse-to-fine architecture or multi-scale architecture were applied~\cite{eigen2014depth,eigen2015predicting,denton2015deep,mathieu2015deep,dosovitskiy2015flownet}.
	However, not all multi-scale CNNs are designed to produce optimal results, similarly to ~\cite{schuler2016learning}. In depth estimation, optical flow estimation, etc., networks usually produce outputs having smaller resolution compared to input image resolution~\cite{eigen2014depth,eigen2015predicting,dosovitskiy2015flownet}. These methods have difficulties in handling long-range dependency even if multi-scale architecture is used.
	
	Therefore, we make a multi-scale architecture that preserves fine-grained detail information as well as long-range dependency from coarser scales. Furthermore, we make sure intermediate level networks help the final stage in an explicit way by training network with multi-scale losses.

	\subsection{Kernel-Free Learning for Dynamic Scene Deblurring}
	
	Conventionally, it was essential to find blur kernel before estimating latent image. CNN based methods were no exception~\cite{schuler2016learning,sun2015learning}. However, estimating kernel involves several problems.
	First, assuming simple kernel convolution cannot model several challenging cases such as occluded regions or depth variations. Second, kernel estimation process is subtle and sensitive to noise and saturation, unless blur model is carefully designed. Furthermore, incorrectly estimated kernels give rise to artifacts in latent images. Third, finding spatially varying kernel for every pixel in dynamic scene requires a huge amount of memory and computation.
	
	Therefore, we adopt kernel-free methods in both blur dataset generation and latent image estimation.
	In blurry image generation, we follow to approximate camera imaging process, rather than assuming specific motions, instead of finding or designing complex blur kernel. We capture successive sharp frames and integrate to simulate blurring process. The detailed procedure is described in section ~\ref{sec:section_dataset}.
	Note that our dataset is composed of blurry and sharp image pairs only, and that the local kernel information is implicitly embedded in it.
	In Fig.~\ref{fig_blur_comp}, our kernel-free blurry image is compared with a conventional synthesized image with uniform blur kernel. Notably, the blur image generated by our method exhibits realistic and spatially varying blurs caused by the moving person and the static background, while the blur image synthesized by conventional method does not.
	For latent image estimation, we do not assume blur sources and train the model solely on our blurry and sharp image pairs. Thus, our proposed method does not suffer from kernel-related problems in deblurring.
	
	\section{Blur Dataset}
	\label{sec:section_dataset}
	
	Instead of modeling a kernel to convolve on a sharp image, we choose to record the sharp information to be integrated over time for blur image generation. As camera sensor receives light during the exposure, sharp image stimulation at every time is accumulated, generating blurry image~\cite{hirsch2011fast}. The integrated signal is then transformed into pixel value by nonlinear CRF (Camera Response Function). Thus, the process could be approximated by accumulating signals from high-speed video frames.
	
	Blur accumulation process can be modeled as follows.
	
	\begin{equation}
	B = g\left(\frac{1}{T} \int_{t=0}^{T}S(t) dt\right) \simeq g\left(\frac{1}{M} \sum_{i=0}^{M-1} S[i]\right),
	\label{equ_blur_generation}
	\end{equation}
	
	where $T$ and $S(t)$  denote the exposure time and the sensor signal of a sharp image at time $t$, respectively. Similarly, $M$, $S[i]$ are the number of sampled frames and the $i$-th sharp frame signal captured during the exposure time, respectively. $g$ is the CRF that maps a sharp latent signal $S(t)$ into an observed image $\hat{S}(t)$ such that $\hat{S}(t) = g(S(t))$, or $\hat{S}[i] = g(S[i])$.
	In practice, we only have observed video frames while the original signal and the CRF is unknown.
	
	It is known that non-uniform deblurring becomes significantly difficult when nonlinear CRF is involved, and nonlinearity should be taken into account. However, currently, there are no CRF estimation techniques available for an image with spatially varying blur~\cite{tai2013nonlinear}. When the ground truth CRF is not given, a common practical method is to approximate CRF as a gamma curve with $\gamma = 2.2$ as follows, since it is known as an aproximated average of known CRFs~\cite{tai2013nonlinear}.
	
	\begin{equation}
	g(x) = x^{1/\gamma}.
	\end{equation}
	
	Thus, by correcting the gamma function, we obtain the latent frame signal $S[i]$  from the observed image $\hat{S}[i]$ by $S[i] = g^{-1}(\hat{S}[i])$, and then synthesize the corresponding blur image $B$ by using (\ref{equ_blur_generation}).
	
	We used GOPRO4 Hero Black camera to generate our dataset. We took 240 fps videos with GOPRO camera and then averaged varying number (7 - 13) of successive latent frames to produce blurs of different strengths. For example, averaging 15 frames simulates a photo taken at 1/16 shutter speed, while corresponding sharp image shutter speed is 1/240.
	Notably, the sharp latent image corresponding to each blurry one is defined as the mid-frame among the sharp frames that are used to make the blurry image.
	Finally, our dataset is composed of 3214 pairs of blurry and sharp images at 1280x720 resolution.
	The proposed GOPRO dataset is publicly available on our website~\footnote{\url{https://github.com/SeungjunNah/DeepDeblur_release}}.
	
	\section{Proposed Method}
	
	In our model, finer scale image deblurring is aided by coarser scale features. To exploit coarse and middle level information while preserving fine level information at the same time, input and output to our network take the form of Gaussian pyramids. Note that most of other coarse-to-fine networks take a single image as input and output.
	
	\subsection{Model Architecture}
	
	\begin{figure}[t]
		\begin{center}
			\includegraphics[width=\linewidth]{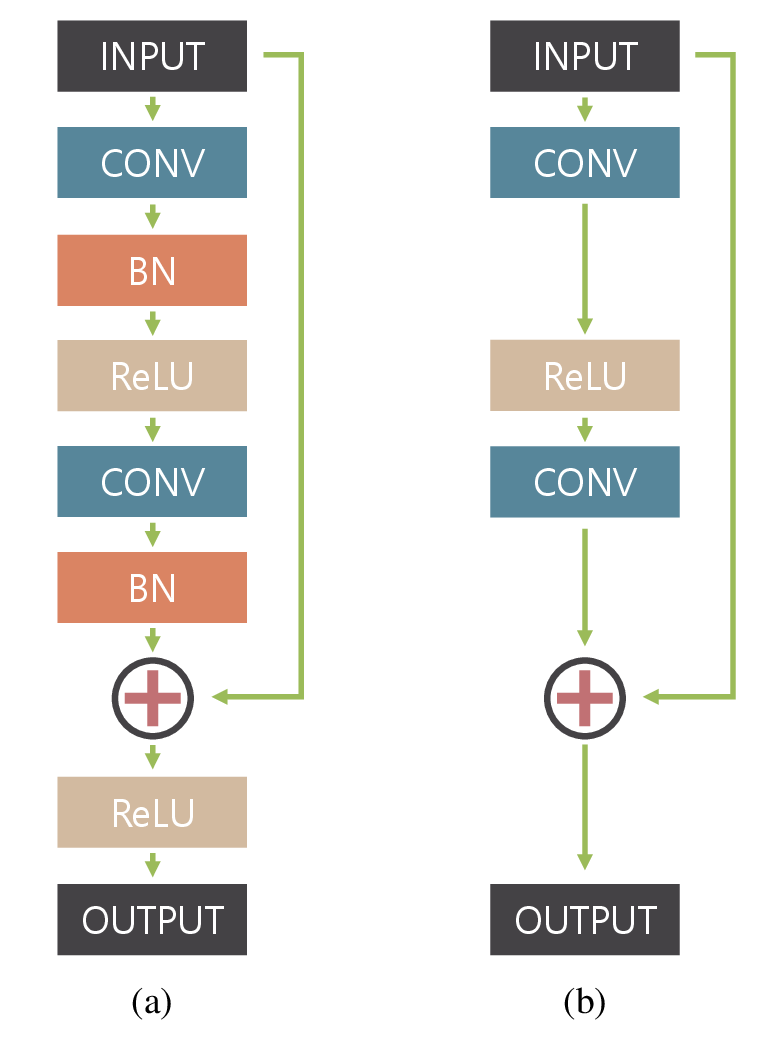}
		\end{center}
		\caption{(a) Original residual network building block. (b) Modified building block of our network.
			We did not use batch normalization layers since we trained model with mini-batch of size 2, which is smaller than usual for batch normalization. We found removing rectified linear unit just before the block output is beneficial in terms of performance empirically.
		}
		\label{fig_resblock}
	\end{figure}
	
	In addition to the multi-scale architecture, we employ a slightly modified version of residual network structure~\cite{he2016deep} as a building block of our model. Using residual network structure enables deeper architecture compared to a plain CNN. Also, as blurry and sharp image pairs are similar in values, it is efficient to let parameters learn the difference only.
	We found that removing the rectified linear unit after the shortcut connection of the original residual building block boosts the convergence speed at training time. We denote the modified building block as ResBlock. The original and our modified building block are compared in Fig.~\ref{fig_resblock}.
	
\begin{figure*}[t]
		\begin{center}
			\includegraphics[width=\linewidth]{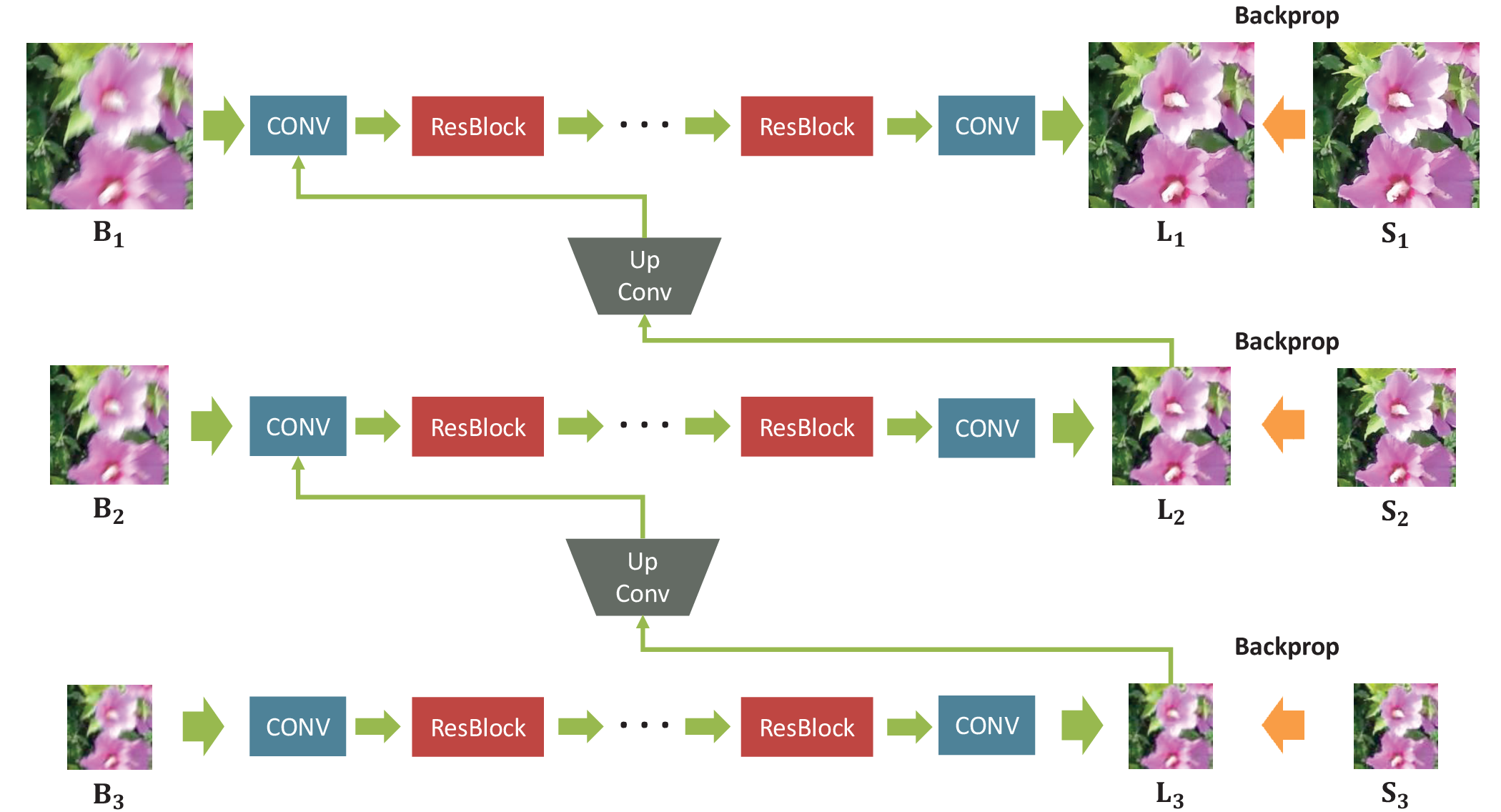}
		\end{center}
		\caption{Multi-scale network architecture. $B_k$, $L_k$, $S_k$ denote blurry and latent, and ground truth sharp images, respectively. Subscript $k$ denotes $k$-th scale level in the Gaussian pyramid, which is downsampled to $1/2^k$ scale. 
			Our model takes a blurry image pyramid as the input and outputs an estimated latent image pyramid. Every intermediate scale output is trained to be sharp. At test time, original scale image is chosen as the final result.}
		\label{fig_architecture}
	\end{figure*}

	By stacking enough number of convolution layers with ResBlocks, the receptive field at each scale is expanded. Details are described in the following paragraphs. For sake of consistency, we define scale levels in the order of decreasing resolution (i.e. level 1 for finest scale). Unless denoted otherwise, we use total $K=3$ scales. At training time, we set the resolution of the input and output Gaussian pyramid patches to be $\{256\times256, 128\times128, 64\times64\}$. The scale ratio between consecutive scales is 0.5. For all convolution layers, we set the filter size to be $5\times5$. As our model is fully convolutional, at test time, the patch size may vary as the GPU memory allows. The overall architecture is shown in Fig.~\ref{fig_architecture}.
	
	\subsubsection*{Coarsest level network}
	
	At the front of the network locates the coarsest level network. The first convolution layer transforms 1/4 resolution, $64\times64$ size image into 64 feature maps. Then, 19 ResBlocks are stacked followed by last convolution layer that transforms the feature map into input dimension. Every convolution layer preserves resolution with zero padding. In total, there are 40 convolution layers. The number of convolution layers at each scale level is determined so that total model should have 120 convolution layers. Thus, the coarsest level network has receptive field large enough to cover the whole patch.
	At the end of the stage, the coarsest level latent sharp image is generated. Moreover, information from the coarsest level output is delivered to the next stage where finer scale network is. To convert a coarsest output to fit the input size of the next finer scale, the output patch passes an upconvolution~\cite{long2015fully} layer, while other multi-scale methods use reshaping~\cite{eigen2014depth} or upsampling~\cite{denton2015deep,eigen2015predicting,mathieu2015deep}. Since the sharp and blurry patches share low-frequency information, learning suitable feature with upconvolution helps to remove redundancy.
	In our experiment, using upconvolution showed better performance than upsampling.
	Then, the upconvolution feature is concatenated with the finer scale blurry patch as an input.

	\subsubsection*{Finer level network}
	
	Finer level networks basically have the same structure as in the coarsest level network. However, the first convolution layer takes the sharp feature from the previous stage as well as its own blurry input image, in a concatenated form. Every convolution filter size is $5\times5$ with the same number of feature maps as in the coarsest level. Except for the last finest scale, there is an upconvolution layer before the next stage. At the finest scale, the original resolution sharp image is restored.
	
	\subsection{Training}
	\label{section_training}
	
	Our model is trained on the proposed GOPRO dataset. Among 3214 pairs, 2103 pairs were used for training and remainings were used for the test.
	To prevent our network from overfitting, several data augmentation techniques are involved. In terms of geometric transformations, patches are randomly flipped horizontally and vertically, rotated by 90 degrees. For color, RGB channels are randomly permuted.
	To take image degradations into account, saturation in HSV colorspace is multiplied by a random number within $[0.5, 1.5]$.
	Also, Gaussian random noise is added to blurry images. To make our network be robust against different strengths of noise, standard deviation of noise is also randomly sampled from Gaussian distribution, $N(0, (2/255)^2)$.
	Then, value outside [0, 1] is clipped. Finally, 0.5 is subtracted to set input and output value range zero-centered, having range [-0.5, 0.5].
	
	In optimizing the network parameters, we trained the model in a combination of two losses, multi-scale content loss and adversarial loss.
	
	\subsubsection*{Multi-scale content loss}
	
	Basically, the coarse-to-fine approach desires that every intermediate output becomes the sharp image of the corresponding scale. Thus, we train our network so that intermediate outputs should form a Gaussian pyramid of sharp images. MSE criterion is applied to every level of pyramids.
	Hence, the loss function is defined as follows:
	
	\begin{equation}
	\mathcal{L}_{cont} = \frac{1}{2 K} \sum_{k=1}^{K} \frac{1}{c_k w_k h_k }\|L_k-S_k\|^2
	\label{equ_loss},
	\end{equation}
	where $L_k, S_k$ denote the model output and ground truth image at scale level $k$, respectively. The loss at each scale is normalized by the number of channels $c_k$, width $w_k$, and the height $h_k$ (i.e. the total number of elements).

	\subsubsection*{Adversarial loss}
	
	Recently, adversarial networks are reported to generate sharp realistic images~\cite{goodfellow2014generative, denton2015deep, radford2015unsupervised}. Following the architecture introduced in ~\cite{radford2015unsupervised}, we build discriminator as in Table~\ref{table_discriminator}. Discriminator takes the output of the finest scale or the ground truth sharp image as input and classifies if it is deblurred image or sharp image.
	
	The adversarial loss is defined as follows.
	
	\begin{equation}
	\begin{multlined}
	\mathcal{L}_{adv} = \mathop{\mathbb{E}}_{S\sim p_{sharp}(S)} [\log D(S)] + \\
	~~~~~~~~~~~\mathop{\mathbb{E}}_{{B}\sim p_{blurry}(B)} [\log (1-D(G(B)))],\\
	\end{multlined}
	\label{adversarial_loss}
	\end{equation}
	
	where \textit{G} and \textit{D} denote the generator, that is our multi-scale deblurring network in Fig.~\ref{fig_architecture} and the discriminator (classifier), respectively. When training, \textit{G} tries to minimize the adversarial loss while \textit{D} tries to maximize it.
	
	\begin{table}[h]
		\renewcommand{\arraystretch}{1.3}
		\begin{center}
			\begin{tabular}{|c||c|c|c|}
				\hline
				
				\ \textbf{\#} & \textbf{Layer} & \textbf{Weight dimension} & \textbf{Stride}\\
				\hline
				\ 1 & conv & $32\times3\times5\times5$ & 2\\
				\ 2 & conv & $64\times32\times5\times5$ & 1\\
				\ 3 & conv & $64\times64\times5\times5$ & 2\\
				\ 4 & conv & $128\times64\times5\times5$ & 1\\
				\ 5 & conv & $128\times128\times5\times5$ & 4\\
				\ 6 & conv & $256\times128\times5\times5$ & 1\\
				\ 7 & conv & $256\times256\times5\times5$ & 4\\
				\ 8 & conv & $512\times256\times5\times5$ & 1\\
				\ 9 & conv & $512\times512\times4\times4$ & 4\\
				\ 10 & fc & $512\times1\times1\times1$ & -\\
				\ 11 & sigmoid & - & -\\
				
				\hline
			\end{tabular}
		\end{center}
		\caption{Model parameters of the discriminator. Every convolution layers are activated with LeakyReLU layer.}
		\label{table_discriminator}
	\end{table}
	
	Finally, by combining the multi-scale content loss and adversarial loss, the generator network and discriminator network is jointly trained. Thus, our final loss term is 
	
	\begin{equation}
	\mathcal{L}_{total} = \mathcal{L}_{cont} + \lambda \times \mathcal{L}_{adv},
	\label{total_loss}
	\end{equation}
	
	where the weight constant $\lambda = 1\times10^{-4}$.
	
	We used ADAM~\cite{kingma2014adam} optimizer with a mini-batch size 2 for training. The learning rate is adaptively tuned beginning from $5\times 10^{-5}$. After $3\times10^5$ iterations, the learning rate is decreased to 1/10 of the previous learning rate. Total training takes $9\times10^5$ iterations to converge.

	\section{Experimental Results}
	
	We implemented our model with torch7 library. All the following experiments were performed in a desktop with i7-6700K CPU and NVIDIA GTX Titan X (Maxwell) GPU.
	
	\subsection{GOPRO Dataset}
	
\begin{figure*}
		\begin{center}
			\includegraphics[width=\linewidth]{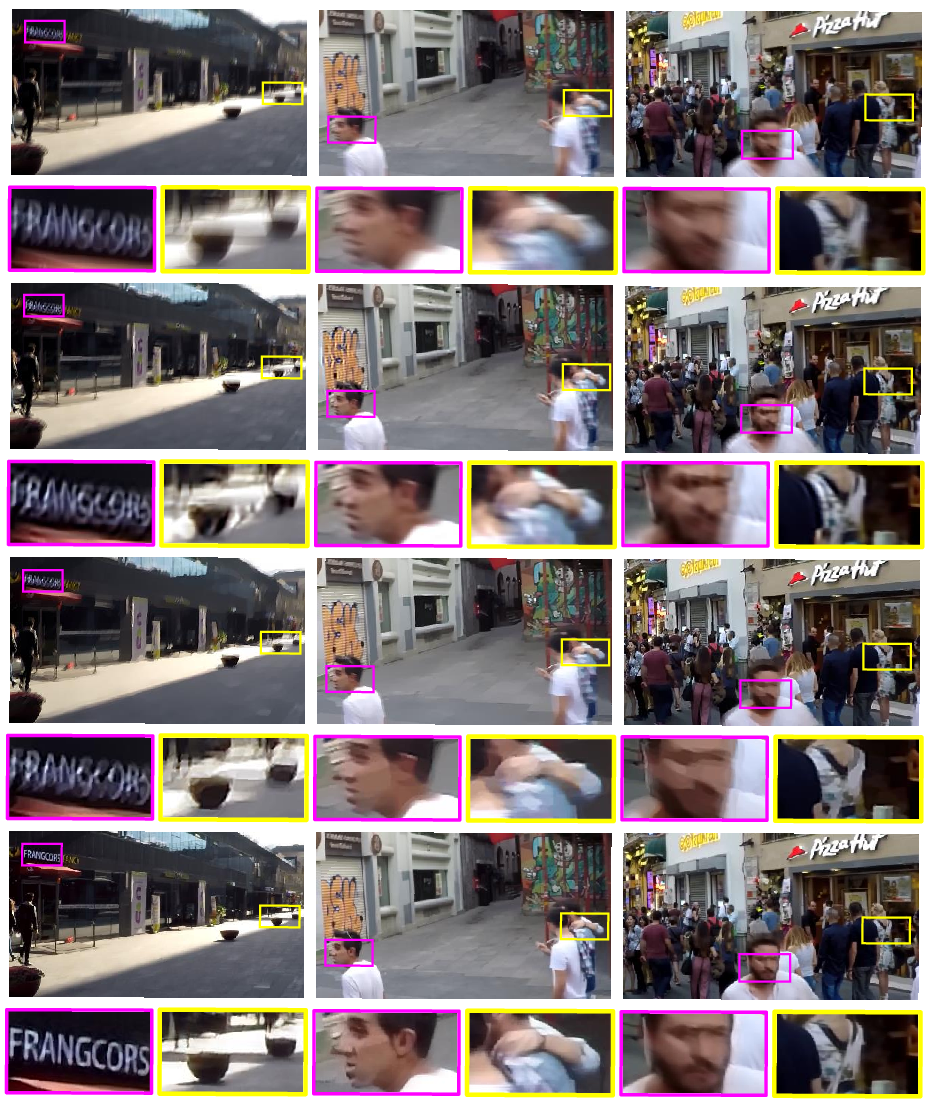}
		\end{center}
		\caption{Test results on the GOPRO dataset. From top to bottom: Blurry images, results of Sun et al.~\cite{sun2015learning}, results of Kim and Lee~\cite{thkim:2014}, and results of the proposed method.}
		\label{fig_result_comparison}
	\end{figure*}

	We evaluate the performance of our model in the proposed GOPRO dataset. Our test dataset consists of 1111 pairs, which is approximately $1/3$ of the total dataset. We compare the results with those of the state-of-the-art methods~\cite{thkim:2014,sun2015learning} in both qualitative and quantitative ways. Our results show significant improvement in terms of image quality.
	Some deblurring results are shown in Fig.~\ref{fig_result_comparison}. We notice from the results of Sun et al.~\cite{sun2015learning}, deblurring is not successful on the regions where blurs are nonlinearly shaped or located at the boundary of motion. Kim and Lee ~\cite{thkim:2014}'s results also fail in cases where strong edges are not found. In contrast, our results are free from those kernel-estimation related problems. Table~\ref{table_evaluation}, shows the quantitative evaluation results of the competing methods and ours with different scale level $k$ in terms of PSNR, SSIM over the test data. Also, the runtime is compared. We observe that our system with $K=2$ produces the best results in terms of both PSNR and SSIM, while $K=3$ is the fastest.
	
	\begin{table}[h]
		\renewcommand{\arraystretch}{1.3}
		\begin{center}
			\begin{tabular}{|c|| c|c|c|c|c|}
				\hline
				
				\multirow{2}{*}{\bfseries Measure} & \multirow{2}{*}{\cite{sun2015learning}} & \multirow{2}{*}{\cite{thkim:2014}} & \multicolumn{3}{c|}{Ours}\\
				\hhline{~~~---}
				& & & $K=1$ & $K=2$ & $K=3$ \\
				
				\hline
				\ PSNR & 24.64 & 23.64 & 28.93 & 29.23 & 29.08 \\
				\hline
				\ SSIM & 0.8429 & 0.8239 & 0.9100 & 0.9162 & 0.9135\\
				\hline
				\ Runtime & 20 min & 1 hr & 7.21 s & 4.33 s & 3.09 s \\
				
				\hline
			\end{tabular}
		\end{center}
		\caption{Quantitative deblurring performance comparison on the GOPRO dataset. $K$ denotes the scale level.}
		\label{table_evaluation}
	\end{table}
	
	\begin{figure*}[t]
		\begin{center}
			\includegraphics[width=\linewidth]{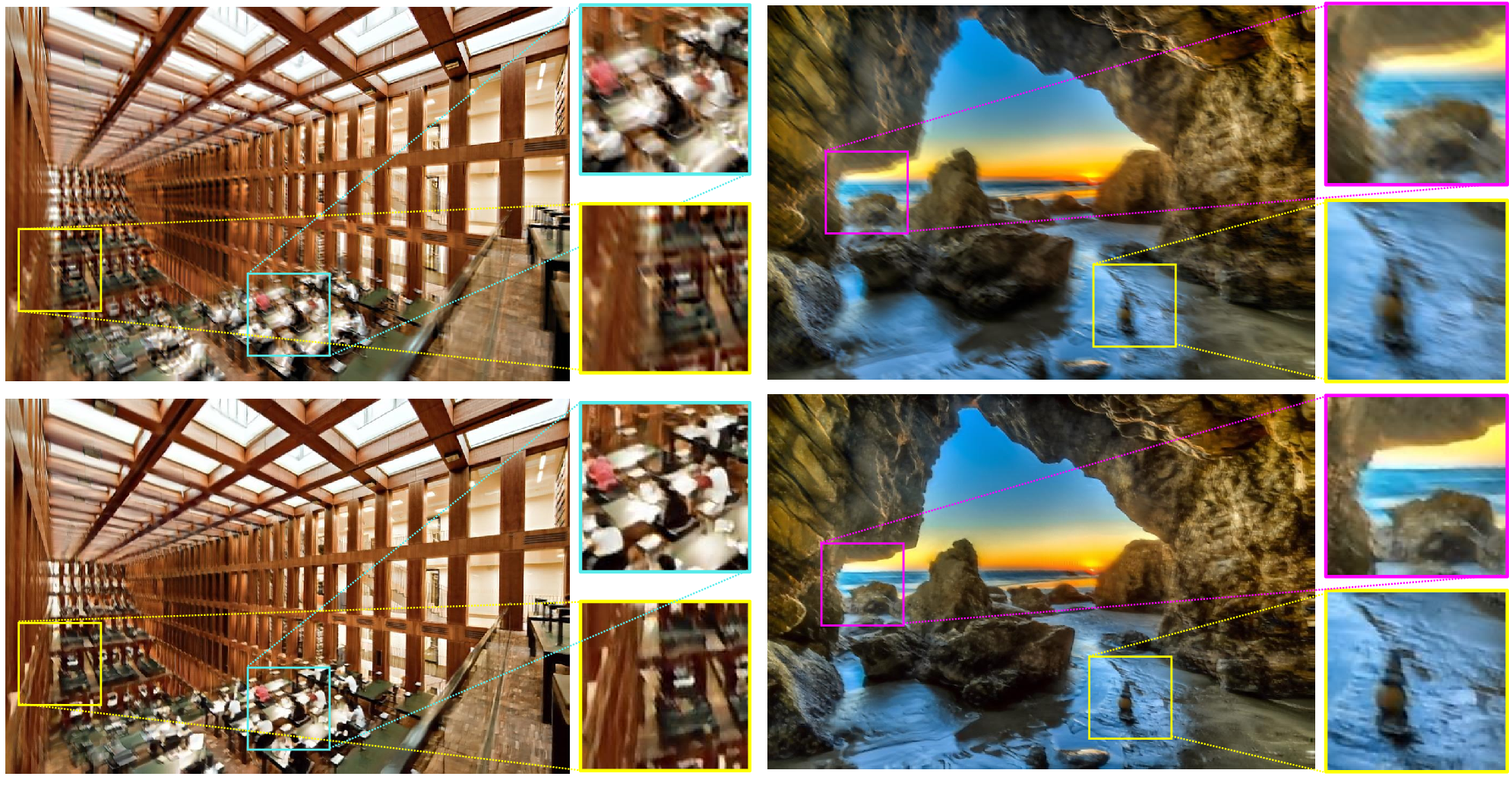}
		\end{center}
		\caption{Deblurring results on the dataset ~\cite{lai2016comparative}. The top row shows results of
			results of Sun et al.~\cite{sun2015learning} and the bottom row shows our results.}
		\label{fig_result_lai}
	\end{figure*}
	
	\subsection{K{\"o}hler Dataset}
	
	K{\"o}hler dataset ~\cite{kohler2012recording} consists of 4 latent images and 12 differently blurred images for each of them. The blurs are caused by replaying recorded 6D camera motion, assuming linear CRF. We report the quantitative results on this dataset in Table~\ref{table_evaluation_kohler}. Our model is trained by setting $g$ as identity function in~(\ref{equ_blur_generation}). We note that our system with $K=3$ produces the best results in PSNR, and the system $K=2$ exhibits the best MSSIM result.
	
	\begin{table}[h]
		\renewcommand{\arraystretch}{1.3}
		\begin{center}
			\begin{tabular}{|c|| c|c|c|c|c|}
				\hline
				
				\multirow{2}{*}{\bfseries Measure} & \multirow{2}{*}{\cite{sun2015learning}} & \multirow{2}{*}{\cite{thkim:2014}} & \multicolumn{3}{c|}{Ours}\\
				\hhline{~~~---}
				& & & $K=1$ & $K=2$ & $K=3$ \\
				
				\hline
				\ PSNR & 25.22 & 24.68 & 25.74 & 26.02 & 26.48 \\
				\hline
				\ MSSIM & 0.7735 & 0.7937 & 0.8042 & 0.8116 & 0.8079 \\
				
				\hline
			\end{tabular}
		\end{center}
		\caption{Quantitative comparison on the K{\"o}hler dataset. The dataset has its own evaluation code, thus we report multi-scale SSIM instead of SSIM.}
		\label{table_evaluation_kohler}
	\end{table}

	\subsection{Dataset of Lai et al.}
	
	\label{experiment_lai}
	
	Lai et al. ~\cite{lai2016comparative} generated synthetic dataset by convolving nonuniform blur kernels and imposing several common degradations. They also recorded 6D camera trajectories to generate blur kernels. However, their blurry images and sharp images are not aligned in the way of our dataset, making simple image quality measures such as PSNR and SSIM less correlated with perceptual quality. Thus, we show qualitative comparisons in Fig.~\ref{fig_result_lai}. Clearly, our results avoid ringing artifacts while preserving details such as wave ripple.

	\section{Conclusion}
	
	In this paper, we proposed a blind deblurring neural network for sharp image estimation. Unlike previous studies, our model avoids problems related to kernel estimation. The proposed model follows a coarse-to-fine approach and is trained in multi-scale space. We also constructed a realistic ground-truth blur dataset, enabling efficient supervised learning and rigorous evaluation. Experimental results show that our approach outperforms the state-of-the-art methods in both qualitative and quantitative ways while being much faster.
	
	\section*{Acknowledgement}
	
	This project is partially funded by Microsoft Research Asia.
	
	{\small
		\bibliographystyle{ieee}

	}

\clearpage
\onecolumn
\setcounter{section}{0}
\setcounter{figure}{0}
\setcounter{table}{0}
\renewcommand\thesection{\Alph{section}}
\renewcommand{\thefigure}{\thesection.\arabic{figure}}
\renewcommand{\thetable}{\thesection.\arabic{table}}

\section{Appendix}

In this appendix, we present more comparative experimental results to demonstrate the effectiveness of our proposed deblurring method.

\subsection{Comparison of loss function}

In section \ref{section_training}, we employed a loss function that combines both the multi-scale content loss (MSE) and the adversarial loss for training our network. We examine the effect of the adversarial loss term quantitatively and qualitatively. The PSNR and SSIM results are shown in table~\ref{table_loss_comp}. From this results, we observe that adding adversarial loss does not increases PSNR, but increase SSIM, which means that it encourages to generate more natural and structure preserving images.

\begin{table}[h]
	\renewcommand{\arraystretch}{1.3}
	\caption{Quantitative deblurring performance comparison of loss used to optimize our model ~($K=3,~ \lambda = 1\times10^{-4}$). Evaluated on the GOPRO test dataset assuming linear CRF.}
	\label{table_loss_comp}
	\begin{center}
		\begin{tabular}{|c|| c|c|}
			\hline
			\ Loss & $\mathcal{L}_{cont}(MSE)$ & $\mathcal{L}_{cont} + \lambda \mathcal{L}_{adv}$ \\
			\hline
			\ PSNR & 28.62 & 28.45 \\
			\hline
			\ SSIM & 0.9094 & 0.9170 \\
			
			\hline
		\end{tabular}
	\end{center}
\end{table}

\begin{figure*}[h]
	\begin{center}
		\includegraphics[width=\linewidth]{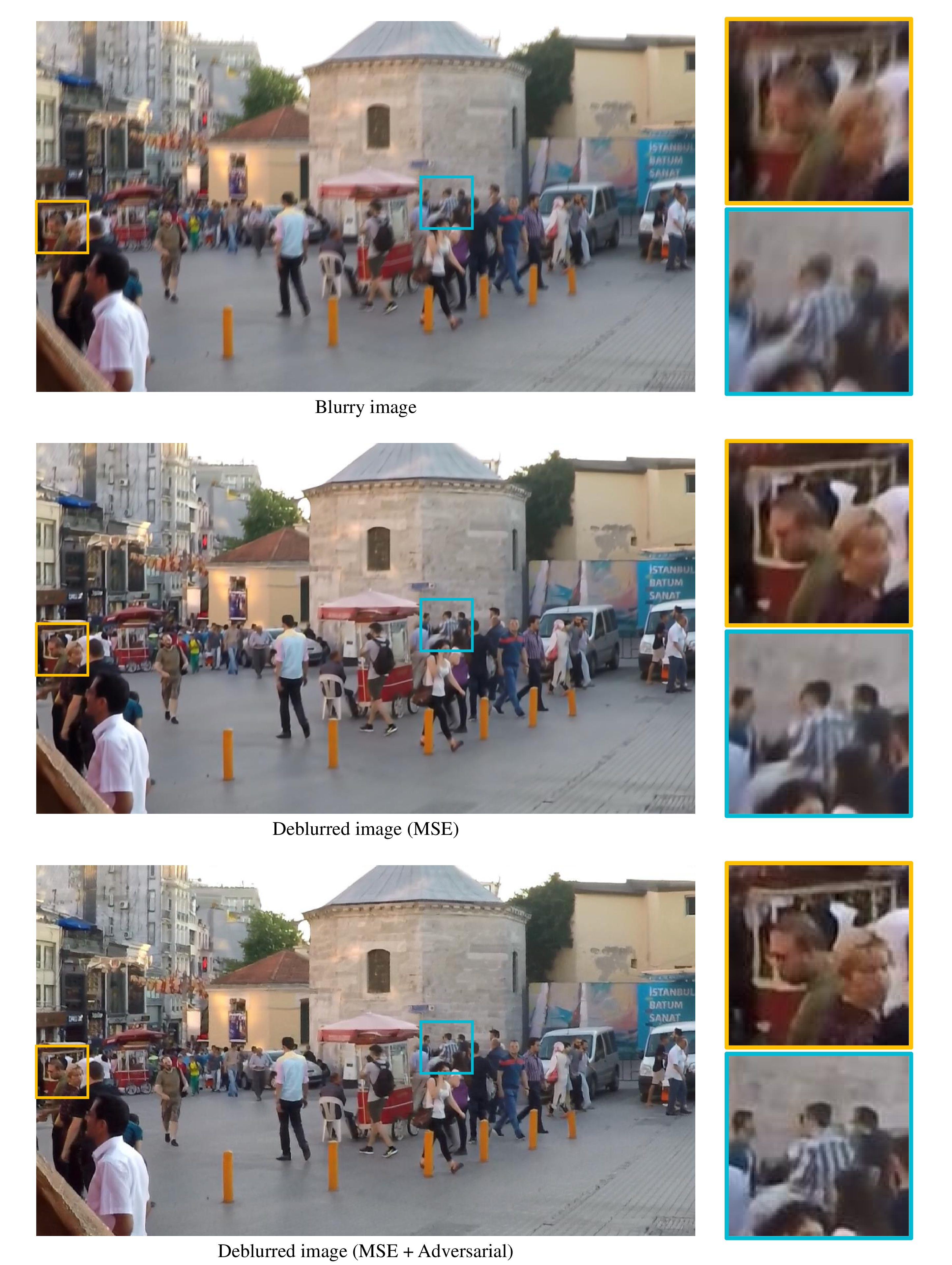}
	\end{center}
	\caption{Visual comparison of results from our model trained with different loss functions. The blurry image is from our proposed dataset.}
	\label{fig_loss_comp1}
\end{figure*}

\begin{figure*}[h]
	\begin{center}
		\includegraphics[width=\linewidth]{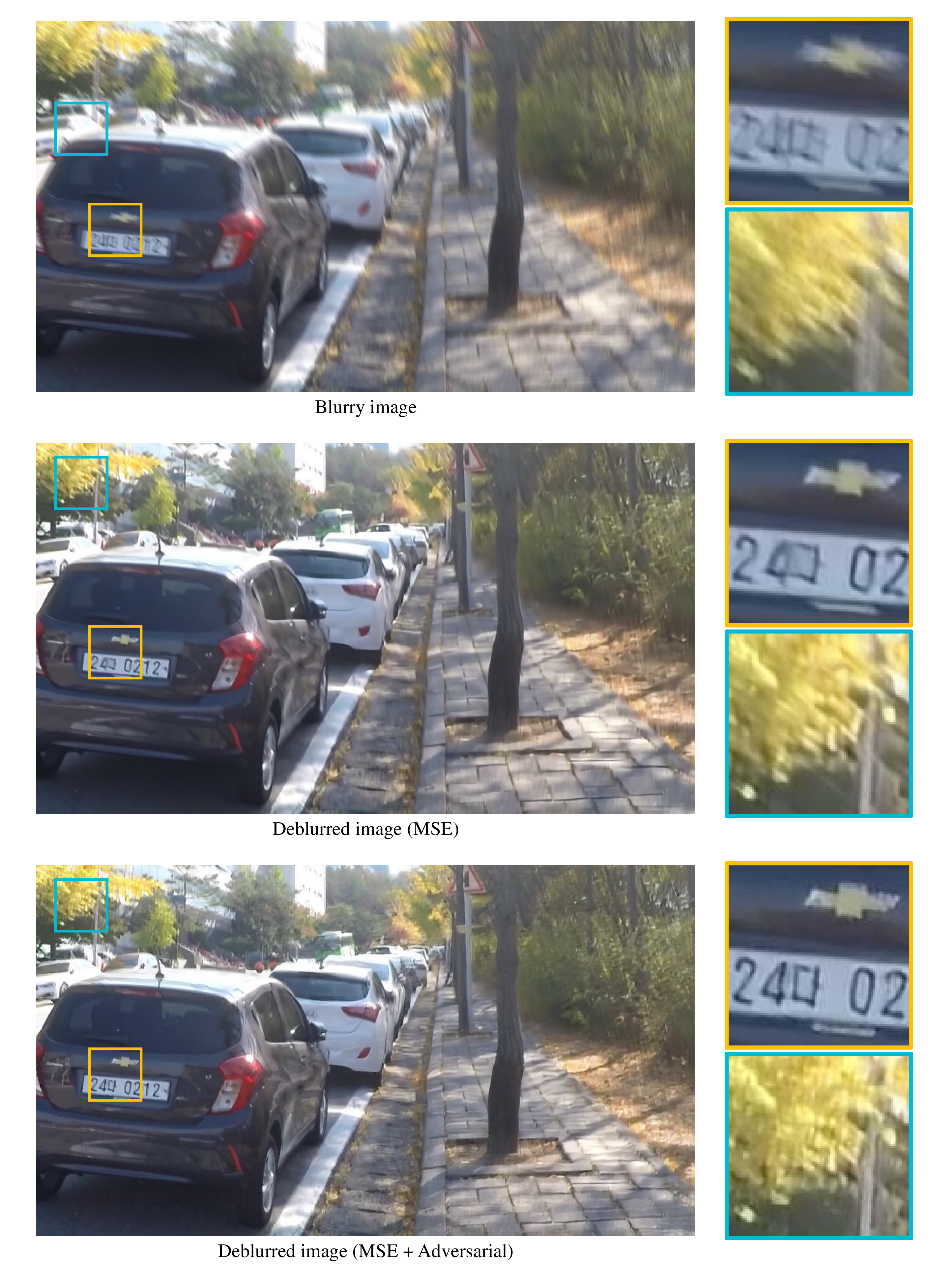}
	\end{center}
	\caption{Visual comparison of results from our model trained with different loss functions. The blurry image is from our proposed dataset.}
	\label{fig_loss_comp2}
\end{figure*}

Fig.~\ref{fig_loss_comp1} and \ref{fig_loss_comp2} show some qualitative comparisons between the results of our network trained with $\mathcal{L}_{cont}$ and $\mathcal{L}_{cont} + \lambda \mathcal{L}_{adv}$.

\clearpage

\subsection{Comparison on GOPRO dataset}

\begin{figure*}[h]
	\begin{center}
		\includegraphics[width=\linewidth]{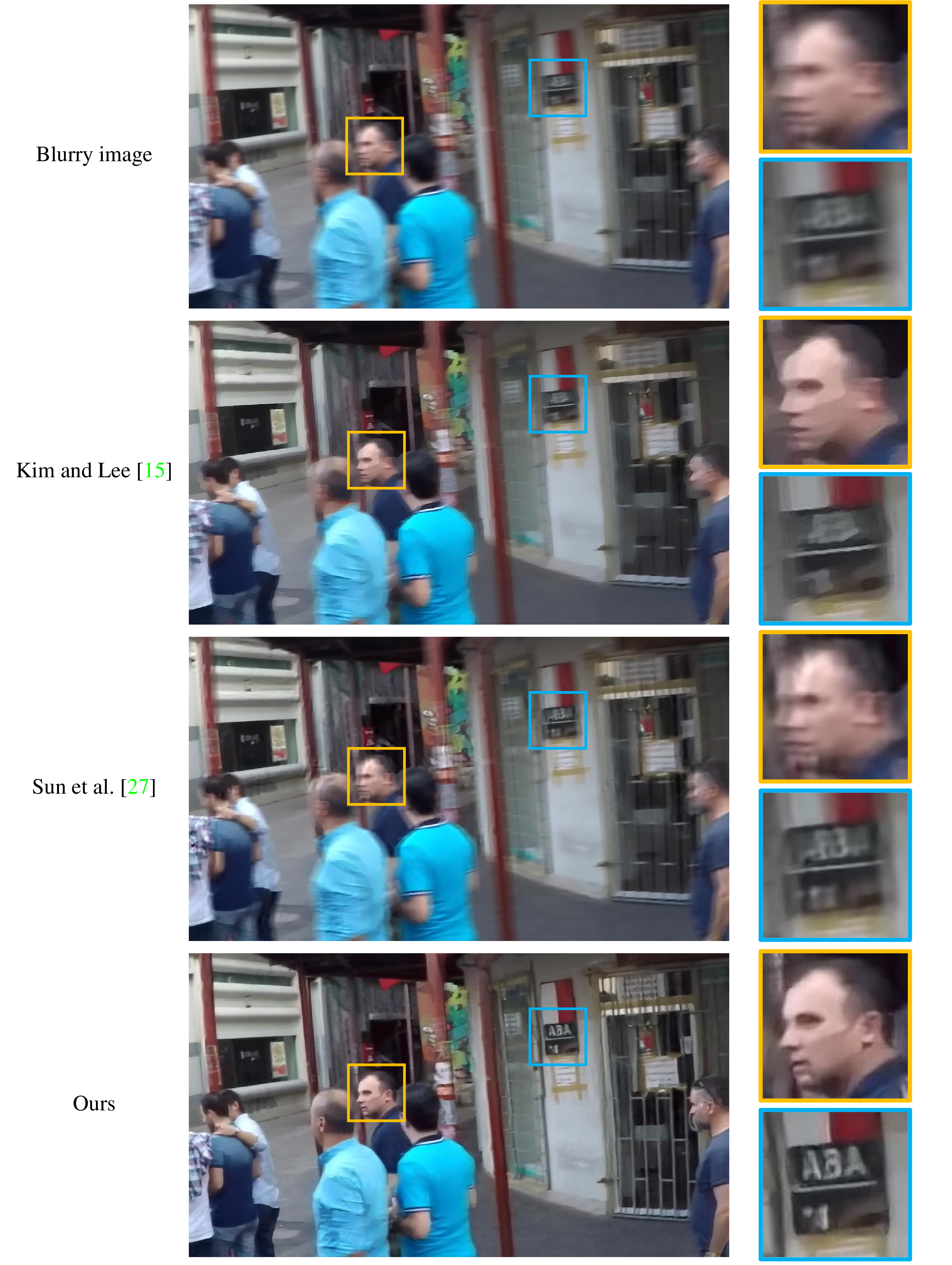}
	\end{center}
	\caption{Visual comparison with other methods. The blurry image is from our proposed dataset.}
	\label{gopro_comp1}
\end{figure*}

\begin{figure*}[h]
	\begin{center}
		\includegraphics[width=\linewidth]{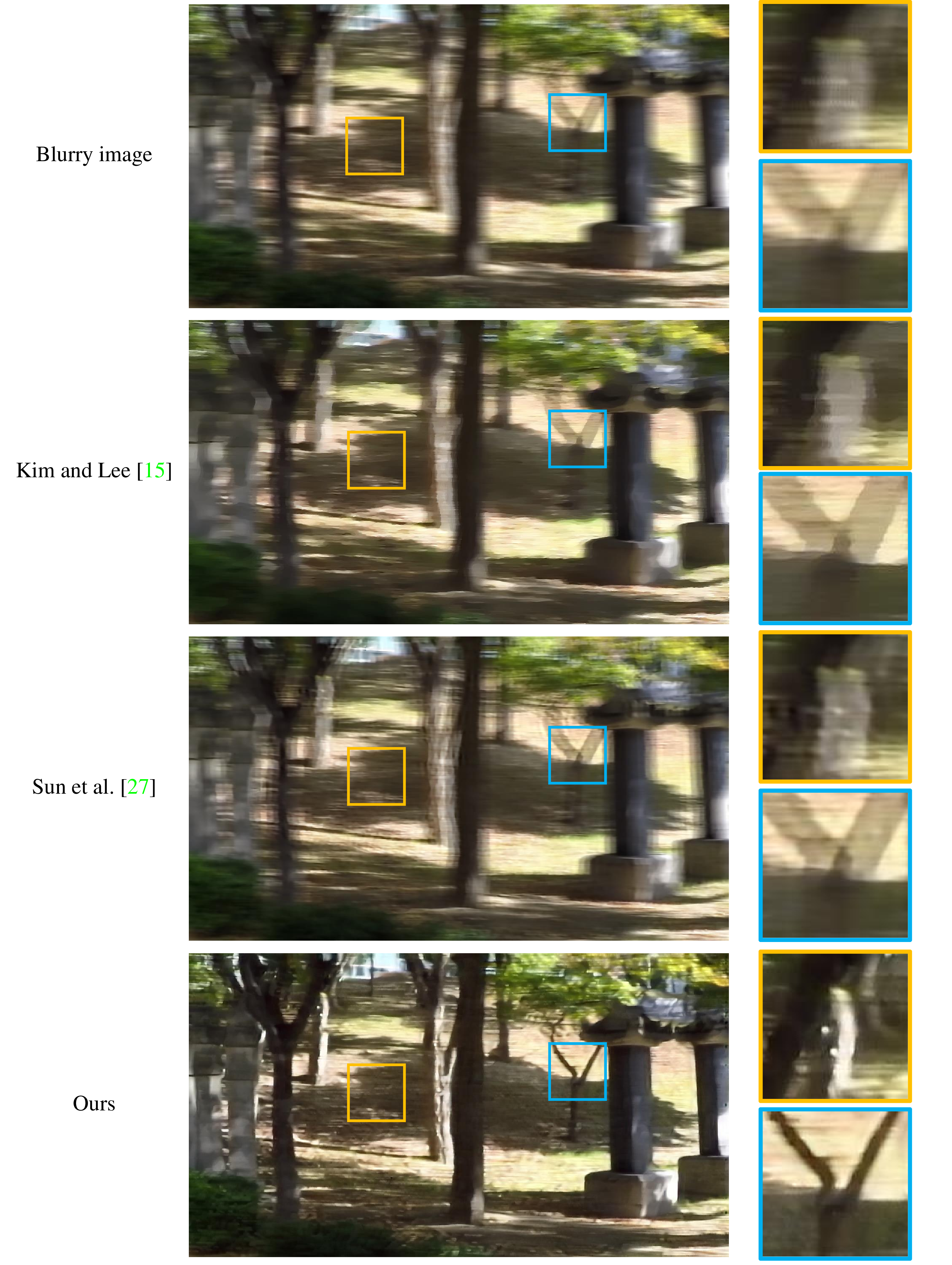}
	\end{center}
	\caption{Visual comparison with other methods. The blurry image is from our proposed dataset.}
	\label{gopro_comp2}
\end{figure*}

We provide qualitative results on our GOPRO test dataset. Fig.~\ref{gopro_comp1} and \ref{gopro_comp2} shows the deblurring results of Kim and Lee~\cite{thkim:2014}, Sun et al.~\cite{sun2015learning}, and ours.

\clearpage

\subsection{Comparison on Lai et al.~\cite{lai2016comparative} dataset}

\begin{figure*}[h]
	\begin{center}
		\includegraphics[width=\linewidth]{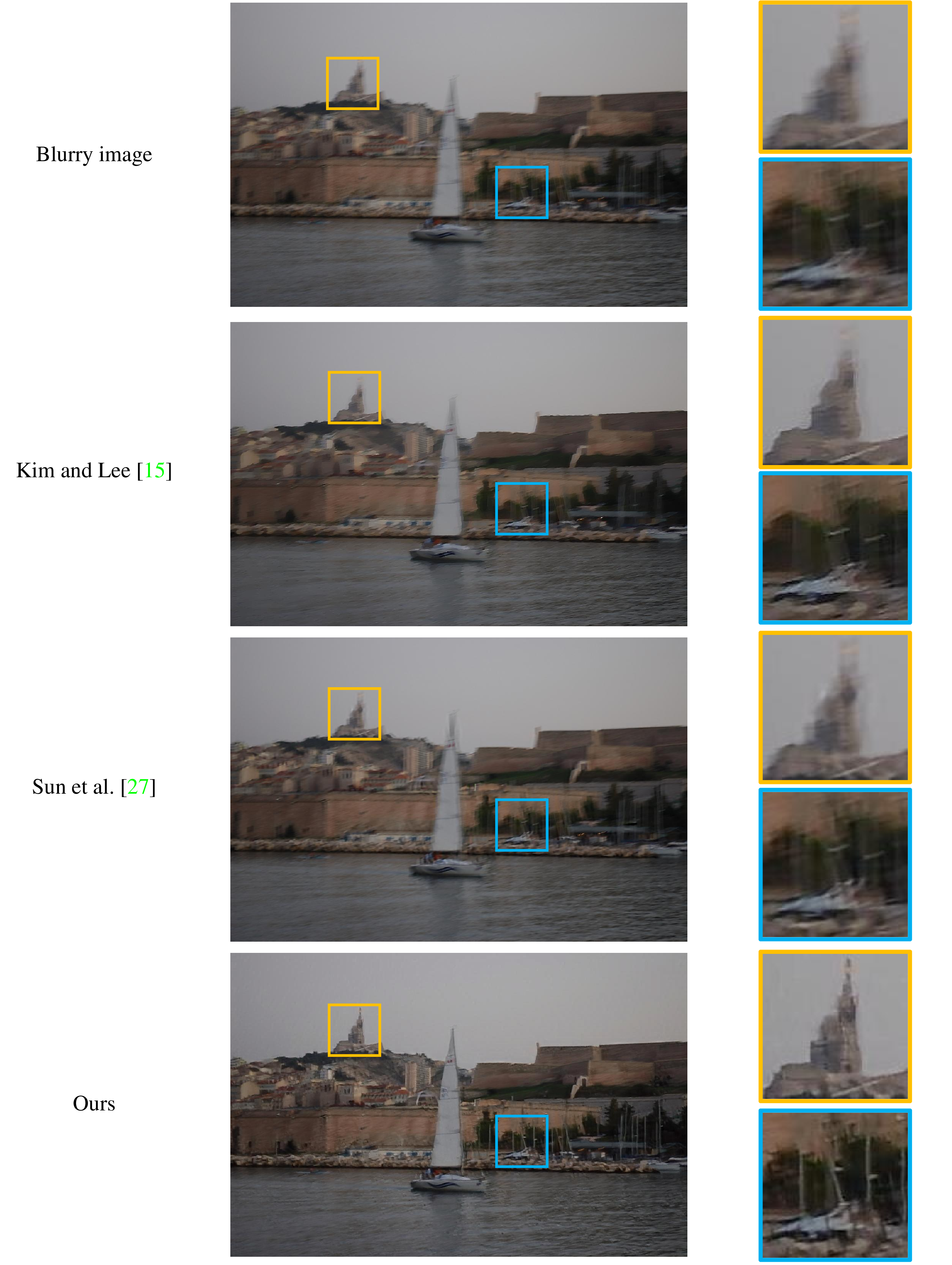}
	\end{center}
	\caption{Visual comparison with other methods. The blurry image is a real image in Lai et al.~\cite{laicomparative}}
	\label{lai_comp1}
\end{figure*}

\begin{figure*}[h]
	\begin{center}
		\includegraphics[width=\linewidth]{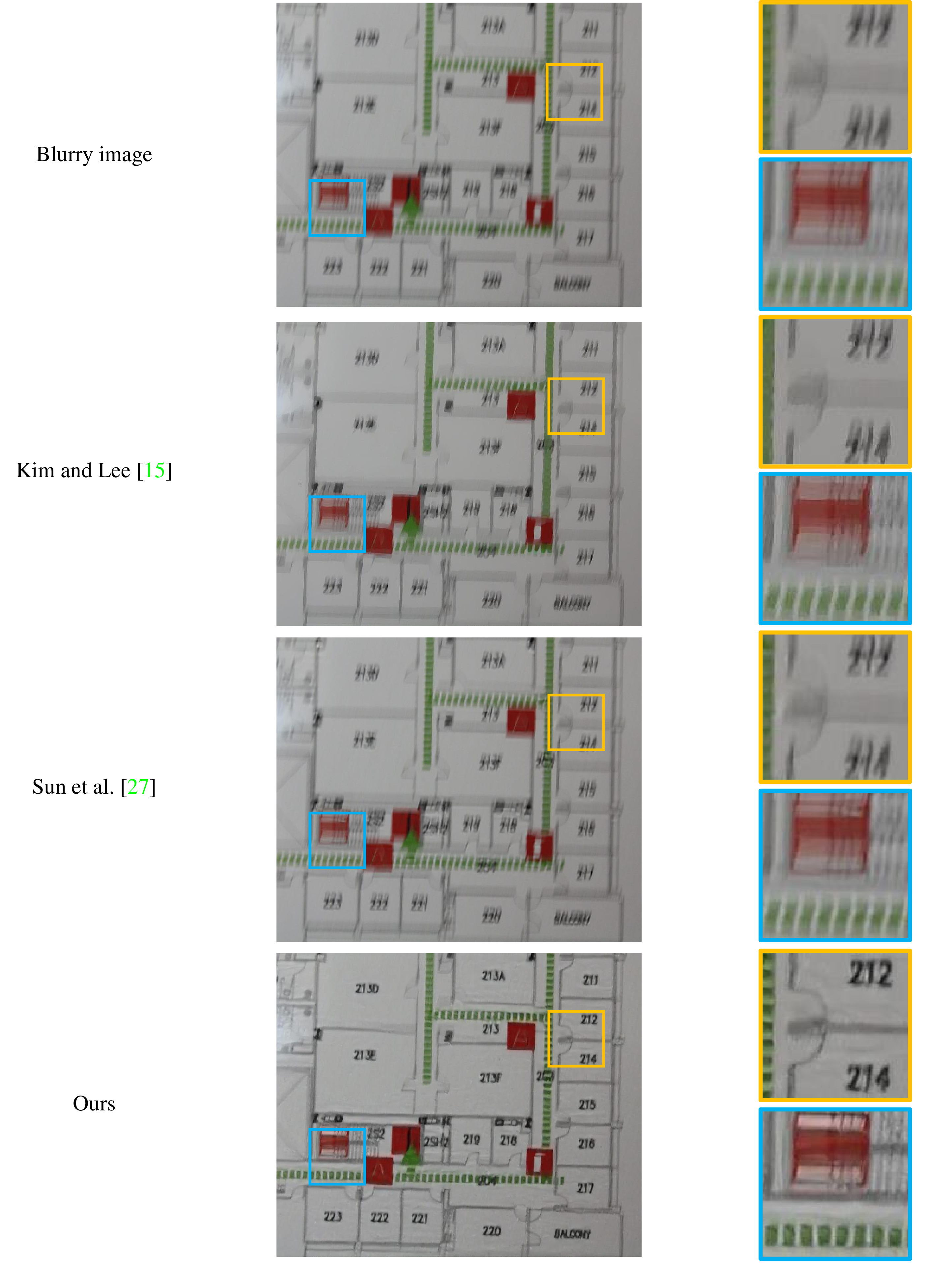}
	\end{center}
	\caption{Visual comparison with other methods. The blurry image is a real image in Lai et al.~\cite{laicomparative}}
	\label{lai_comp2}
\end{figure*}

We provide qualitative results on the dataset of Lai et al. \cite{laicomparative}. The Lai et al. dataset is composed of synthetic and real blurry images, and we showed the deblurring result of a synthetically generated blurry image in section \ref{experiment_lai}. We present the qualitative deblurring results of competing methods on real images in Fig.~\ref{lai_comp1} and \ref{lai_comp2}.

\clearpage

\subsection{Comparison on real dynamic scenes}

\begin{figure*}[h]
	\begin{center}
		\includegraphics[width=\linewidth]{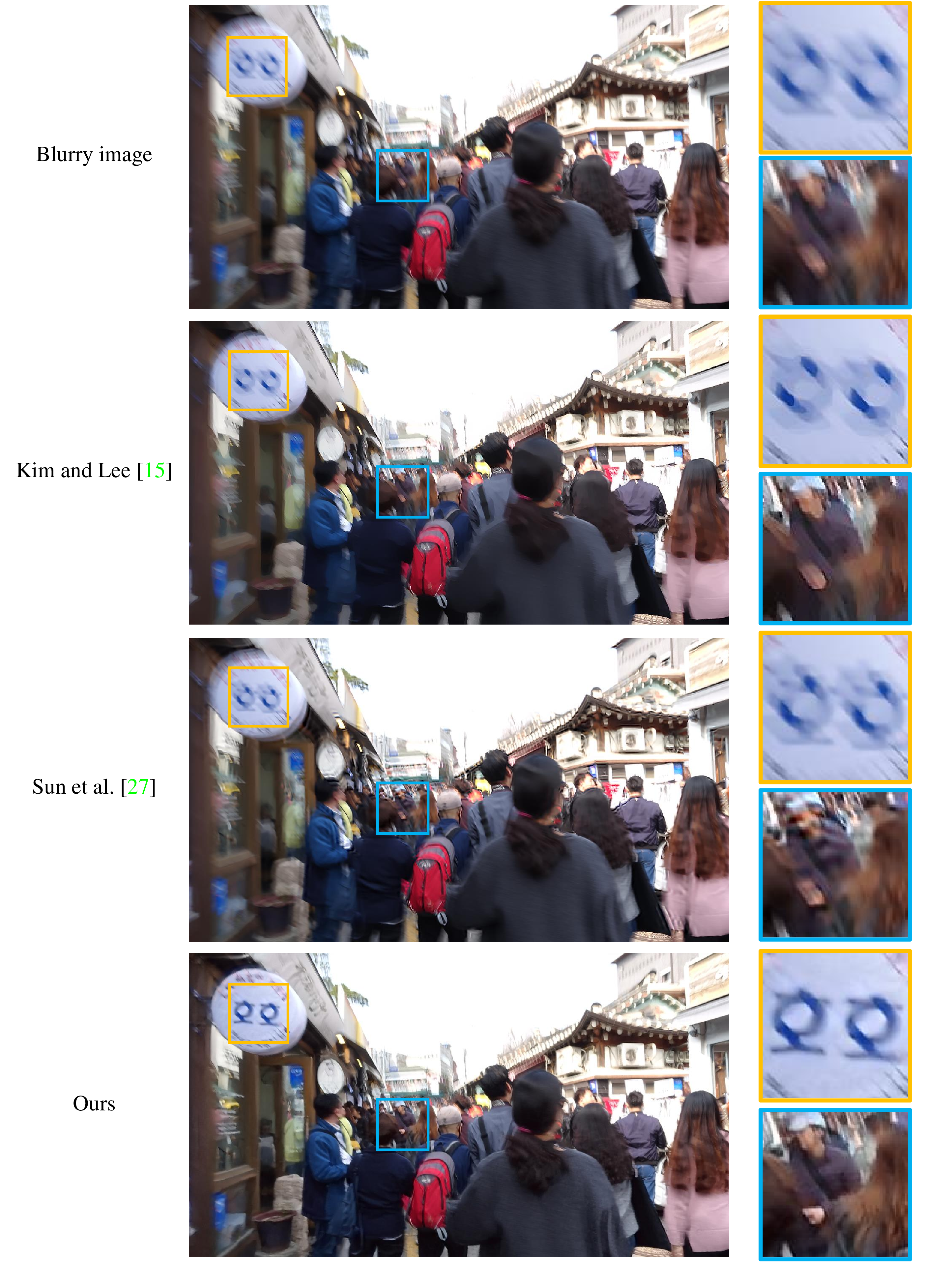}
	\end{center}
	\caption{Visual comparison with other methods. The blurry image is a real dynamic scene.}
	\label{dynamic_comp1}
\end{figure*}

\begin{figure*}[h]
	\begin{center}
		\includegraphics[width=\linewidth]{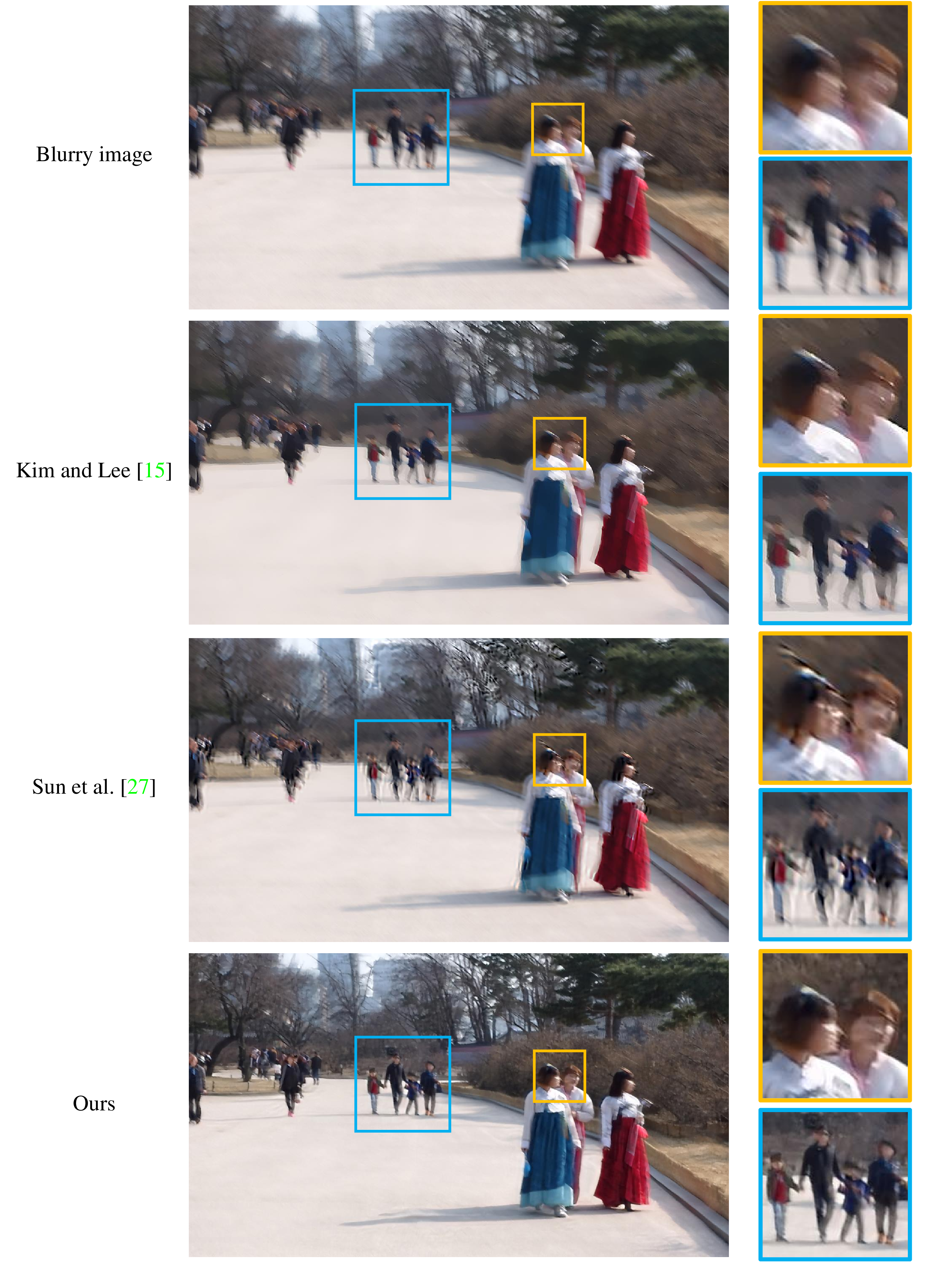}
	\end{center}
	\caption{Visual comparison with other methods. The blurry image is a real dynamic scene.}
	\label{dynamic_comp2}
\end{figure*}

Finally, we further present deblurring results on real dynamic scenes. The blurry scenes are captured by a SONY RX100 M4 camera. The qualitative deblurring results of Kim and Lee~\cite{thkim:2014}, Sun et al.~\cite{sun2015learning} and ours are compared in Fig.~\ref{dynamic_comp1} and \ref{dynamic_comp2}.

\clearpage

\end{document}